# Locomoción bípeda mediante técnicas geométricas

Locomoción Bípeda

*Bipedal locomotion using geometric techniques*

*Bipedal Locomotion*

Antonio Losada González; Dr. Manuel Pérez-Cota
Grupo SI1-GEAC; Departamento de Informática Universidad de Vigo
Ourense, España
alosada@xunta.es ; mpcota@uvigo.es

*Resumen* — **En el presente artículo se pretende describir un algoritmo de caminata bípeda con resolución cinemática inversa basada únicamente en métodos geométricos, de modo que todos los conceptos matemáticos sean explicados desde la base, con el fin de aclarar el porqué de esta solución. Para ello se ha tenido que realizar una labor de simplificación del problema y una labor didáctica de distribución de contenidos. En general, los artículos relacionados con esta temática emplean sistemas matriciales para resolver tanto la cinemática directa como inversa, utilizando técnicas complejas como el desacoplamiento o el cálculo del Jacobiano. Mediante la simplificación del proceso de caminata, se ha propuesto su resolución de modo sencillo empleando únicamente técnicas geométricas.**

*Palabras Clave: Usabilidad; accesibilidad; Inteligencia Artificial; robótica.*

*Abstract* — **The aim of this article is to describe a bipedal walk algorithm with inverse kinematic resolution based solely on geometrical methods, so that all the mathematical concepts are explained from the base, in order to clarify the reason for this solution. For this, it has been necessary to carry out a task of simplifying the problem and a didactic task of content distribution. In general, articles related to this topic used matrix systems to solve both direct and inverse kinematics, using complex techniques such as decoupling or calculating the Jacobian. By simplifying the walking process, its resolution has been proposed in a simple way using only geometric techniques.**

## I. INTRODUCCIÓN

Desde siempre, los seres humanos han querido desarrollar robots a su imagen y semejanza. Esta premisa implicaba la necesidad que éstos tuvieran algo similar a nuestras mismas extremidades. El poseer únicamente dos piernas, implicaba que los mecanismos de control de estabilidad no iban a ser sencillos, dado que resultaría complejo conseguir que el robot se encontrara en todo momento en equilibrio estático. En 1968, Miomir Vukobratovic [2] incorporó el concepto "Zero Moment Point" al mundo de la locomoción robótica, y desde entonces ha formado parte de casi todos los desarrollos de locomoción existentes. En el presente artículo se emplea el concepto ZMP para conseguir una caminata estable, si bien no será necesario introducir la complejidad de su cálculo. Adicionalmente a la incorporación de este concepto y la definición de las fases de una caminata, será necesario realizar la resolución cinemática para determinar los ángulos a aplicar a cada miembro en cada una de las fases de la caminata. En este artículo se pretende explicar de un modo claro, preciso y con formulaciones matemáticas simples, el proceso de locomoción bípeda estable y cómo ésta ha sido utilizada en el desarrollo realizado. Para la resolución de la cinemática inversa se ha empleado el método geométrico, que se encuentra documentado en cientos de artículos relacionado con la resolución de la cadena cinemática inversa de robots de tres grados de libertad, si bien, resulta muy complejo encontrar artículos que resuelvan la cadena cinemática

## II. CINEMÁTICA DE ROBOTS

En la resolución de la cinemática de los robots articulados nos encontramos con dos problemas principales, la cinemática directa y la cinemática inversa.

La cinemática directa resuelve el problema de conocer la posición del efector final una vez conocido el punto de la base del robot y los parámetros del robot, como son las longitudes de todos sus segmentos, y los ángulos de giro de todos sus puntos de libertad, junto el tipo de cada articulación.

La cinemática directa puede resolverse o bien por métodos geométricos o bien por métodos algebraicos

En cuanto a la resolución cinemática inversa, las resoluciones más comunes son las siguientes:

1. Métodos geométricos
2. A partir de las matrices de transformación homogénea
3. Por desacoplamiento cinemática

La mayoría de los métodos de resolución cinemática que podemos ver en artículos o tesis emplean técnicas matriciales, que son aplicables a un número indefinido de elementos para realizar la resolución de las ecuaciones necesarias para llevar a cabo los cálculos cinemáticos y así obtener los ángulos de las articulaciones del robot. Lo que sucede es que, en la mayoría de estos documentos, además de su complejidad, llegan a ser poco educativos, ya que asumen que el lector tiene los conocimientos matemáticos suficientes y comprende el funcionamiento de estos principios de cálculo matricial. Evidentemente los redactores de estos artículos o trabajan directamente en estos campos, o bien han dedicado un precioso tiempo a entender estos conceptos. La intención del presente artículo no es volver a explicar las técnicas de resolución cinemática empleando estos mismos algoritmos, ya que hay cientos de documentos en esta línea en la web, se pretende simplificar y desarrollar estos mismos algoritmos empleando únicamente cálculos geométricos, de modo que se consiga reducir la complejidad matemática empleada.

Para conseguir este objetivo vamos a dividir este trabajo en cuatro apartados principales. En el primer apartado haremos una ligera introducción a la dinámica de funcionamiento del punto de momento cero. Una vez conocida la base de funcionamiento que nos permitirá mantener el equilibrio en todo momento de forma estática pasaremos a estudiar en un segundo apartado cada una de las fases de la caminata bípeda. Finalmente, en este punto ya hemos analizado nuestro problema de forma completa, por lo que en el tercer apartado pasaremos a resolver la cinemática de la caminata bípeda empleando únicamente algoritmos geométricos

DESARROLLO DEL ALGORITMO

### A. Robot bípedo

Lo primero que se va a definir es el tipo del robot al que se puede aplicar estos algoritmos. El algoritmo presentado está diseñado para ser empleado en robots bípedos. Inicialmente ha sido desarrollado para robots con 5 grados de libertad en cada una de sus extremidades inferiores. En una futura ampliación será empleado para robots con capacidad de giro en sus extremidades inferiores.

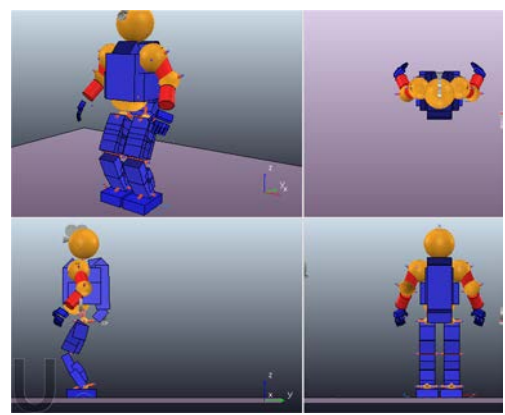

Fig. 1. Representación de la capa física

En la imagen de la Fig. 1 se puede ver la capa física de un modelo estándar de robot bípedo

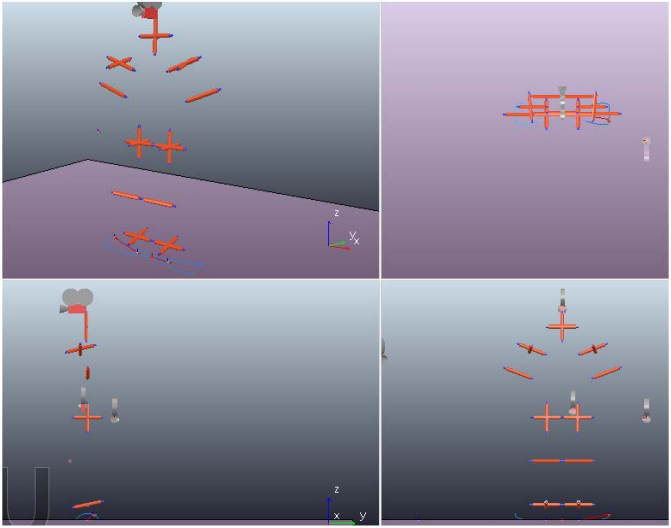

Fig. 2. Grados de libertad

Como se puede ver en las imágenes de la Fig. 2, este robot tiene un total de 20 grados de libertad (reunidos por extremidades) que están marcados por cada uno de los ejes de rotación. Teniendo en cuenta que no tiene ninguno de ellos en las manos, lo sitúa como uno de los sistemas físicos bípedos más complejos. En cuanto al problema de resolución cinemática para conseguir que realice una caminata bípeda, es necesario resolver 10 grados de libertad, es decir, 5 por cada pierna. Este mismo algoritmo podría emplearse para resolver también el ángulo de giro de cada pierna y así conseguir una locomoción más fluida, aunque esto último no es totalmente necesario y no se ha incluido, ya que los robots bípedos más económicos carecen de este eje de rotación.

### B. Zero Moment Point (Punto de momento cero)

Las personas tienen, normalmente, claro que hay muchas formas de andar sin caerse, pero también se tiene claro que hay una técnica entre ellas que se considera la más estable. Esto no implica que sea la mejor, la más rápida, ni la que menor energía consuma, puede ser todo lo contrario, pero resulta la más sencilla de implementar porque, en todo momento, siempre que las velocidades de desplazamiento sean bajas y se pueda ignorar las inercias, el robot se mantiene en un estado de equilibrio estable, lo que quiere decir que se puede parar el movimiento en cualquier momento y el robot no se caerá. Evidentemente esta técnica es la menos parecida al modo de andar que tienen los humanos, pero es la empleada por la inmensa mayoría de los robots bípedos, incluido ASIMO de HONDA (https://www.honda.mx/asimo/), que es una de las representaciones de más alta tecnología en lo referente a robots bípedos. Cabe señalar que ASIMO hace tiempo que es capaz de abandonar su equilibrio estático y es capaz de correr, igual que otros robots que le han sucedido. Evidentemente, si se habla de tecnología punta en este campo, se tienen robots como los de Boston Dynamics (https://www.bostondynamics.com/) que emplean técnicas muchísimo más avanzadas basadas en equilibrio dinámico y retroalimentación de señales y consiguen verdaderas maravillas como las que se pueden ver en muchos de sus videos.

El algoritmo realizado se basa en el concepto ZMP que son las siglas de Zero Moment Point, traducido como Punto de Momento Cero. Para los autores, el momento se puede entender como la fuerza que actúa sobre cualquiera de los puntos del robot en un instante de tiempo. Si se consigue que la resultante de la suma de todas las fuerzas que actúan sobre el robot sea cero, esto implica que no se tendrá ninguna fuerza aplicada al mismo, por lo que el robot será estable y no se moverá. Esto que parece muy complejo, y realmente lo es si se tiene en cuenta todos los factores implicados como aceleraciones, velocidades, inercias, etc., realmente es muy fácil de aplicar si se ignora parte de ellas. En el caso actual de los autores, los movimientos (debidos a la aplicación del robot) van a ser relativamente lentos, lo suficiente para poder ignorar las inercias.

La lógica de ZMP dice que el centro de masa del robot debe estar siempre encima del polígono de soporte. Esta frase que puede sonar extraña es muy sencilla de entender, y para ello se muestran estos conceptos en el robot simulado.

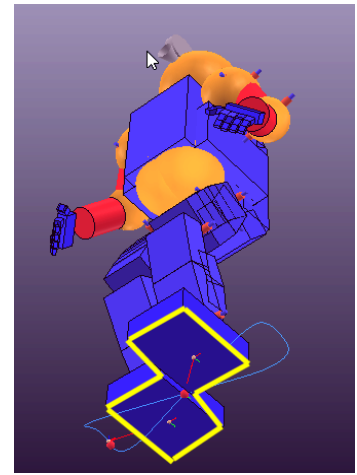

Fig. 3. Polígono de soporte

En la imagen superior se puede ver representado el polígono de soporte, que básicamente cubre toda la superficie de contacto del robot con el suelo.

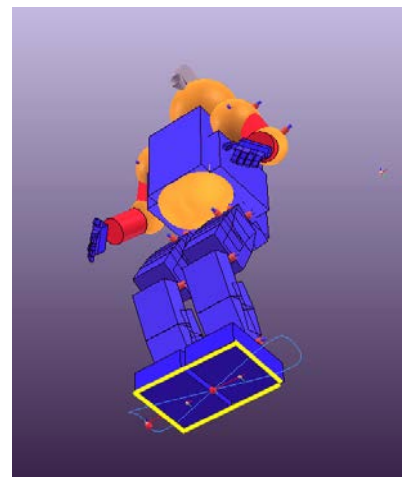

Fig. 4. Polígono de soporte

En las imágenes se puede ver lo que se entiende por el polígono de soporte. El perímetro del polígono de soporte está formado por la unión de los puntos que delimita la planta de los dos pies y el hueco entre ellos. Se debe tener en cuenta que en el polígono de soporte solo intervienen los dos pies si ambos están en contacto con el suelo, si no, se encuentra formado únicamente por la planta del pie que está en contacto con el suelo.

Ahora se explica lo que es el centro de masas. La definición técnica es la siguiente:

> El centro de masas de un sistema discreto o continuo es el punto geométrico que dinámicamente se comporta como si en él estuviera aplicada la resultante de las fuerzas externas al sistema.[8]

El robot se comporta como si toda la masa estuviera concentrada en el centro de masas y el resto del cuerpo no tuviera masa.

En la siguiente imagen, el centro de masas está marcado con el punto central de donde parten las cuatro flechas.

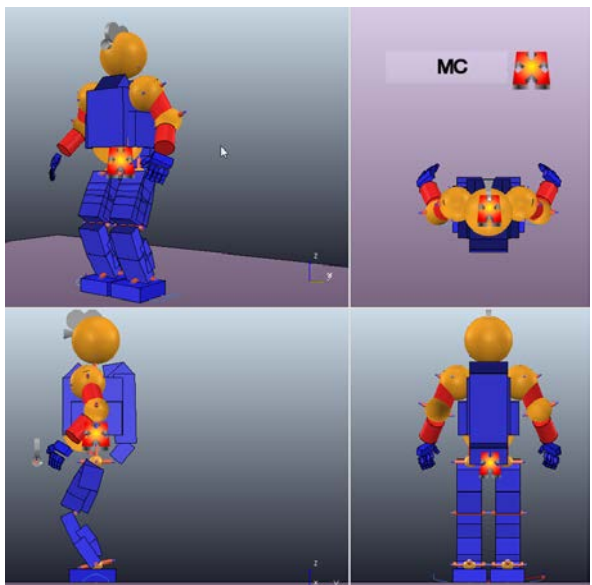

Fig. 5. Centro de masas

Calcular el centro de masas puede resultar muy complejo, pero un criterio que da buenos resultados es colocar el centro de masas en el centro de la cintura del robot. En la imagen superior (Fig. 5) se puede ver el punto escogido para situar el centro de masas.

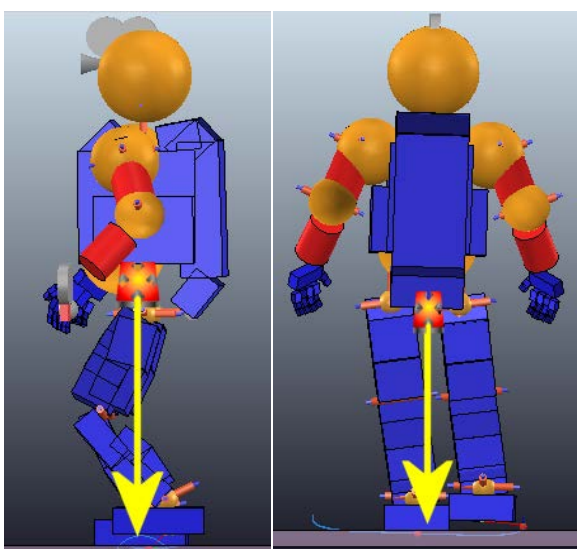

Fig. 6. Movimientos de caminata

En la imagen (Fig. 6) se puede ver uno de los movimientos más característicos de caminatas bípedas realizadas por robots que emplean la técnica ZMP. Este movimiento consiste en desplazarse lateralmente antes de levantar el pie contrario. En la imagen (Fig. 6) se puede observar cómo este desplazamiento permite mantener el centro de masas (cruz roja) encima del polígono de soporte (Ver flecha amarilla).

### C. Fases de la caminata bípeda

El siguiente paso consiste en formular el problema que se pretende solucionar. Para ello se deben analizar las distintas fases de la caminata bípeda para, posteriormente, intentar alcanzar estos movimientos empleando una solución algorítmica.

En las imágenes que acompañan cada una de las fases se ven las posiciones finales de cada fase

#### 1. Posición inicial

En esta primera fase, el robot se encuentra en posición estática. Esta posición es anterior al comienzo de la caminata.

En este punto llama la atención que el robot se encuentra en una postura antinatural, con las piernas flexionadas. Las personas una vez saben andar no tienen esta postura.

Se debe tener en cuenta que se está diseñando un sistema de caminata basado en equilibrio estático. Si se viera en un principio que las piernas están estiradas, en cuanto se iniciara la marcha automáticamente se entraría en una posición de equilibrio dinámico, lo que ocasionaría que el robot comenzara a caer hacia adelante hasta que el pie tocara con el suelo. Con este algoritmo se pretende que todas las fases y posiciones del movimiento sean estables, con lo que el robot puede cesar en su movimiento en cualquier momento y permanecería estable y quieto, salvo por los efectos generados por la inercia de la velocidad del propio movimiento.

Es importante indicar la dificultad de este inicio de movimiento, simplemente, la persona que esté leyendo el artículo, piense en una criatura cuando empieza a dar sus primeros pasos, piense en la forma en que pone el cuerpo y las piernas, por lo que verá muy fácilmente de qué se está hablando en este punto.

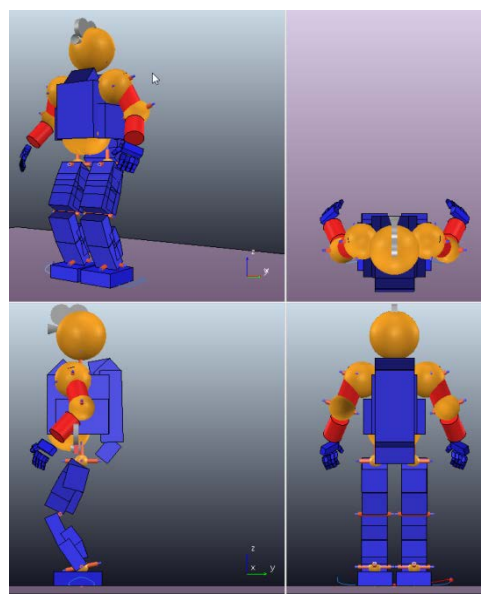

Fig. 7. Robot en posición inicial

Para apreciar con más claridad la dinámica del movimiento del robot, en todas las secuencias se incorporará un gráfico en el que se indican los polígonos de soporte (contacto con el suelo) de los dos pies junto con la posición de la cadera, en la que podremos distinguir las articulaciones laterales y la posición del centro de masas. La imagen representaría en robot visto desde arriba en el que solo vemos los pies y la cadera. El gráfico que representa la posición inicial es el siguiente:

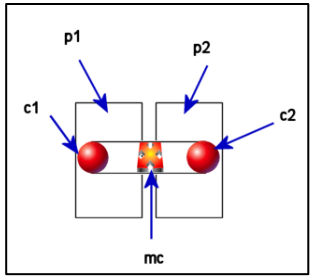

Fig. 7.1. Robot en posición inicial

En la figura se distinguen los siguientes componentes:

p1: pie izquierdo

p2: pie derecho

c1: articulación de la cadera izquierda

c2: articulación de la cadera derecha

mc: centro de masas

### 2. Desplazamiento lateral

En esta fase se desplaza el cuerpo lateralmente. Dado que en la siguiente fase se va a levantar uno de los pies y dejar el robot sobre el otro, es decir, un solo pie. Antes de realizar este movimiento se debe garantizar que el centro de masas se encuentre sobre el polígono de soporte definido por el pie que se va a quedar en contacto con el suelo.

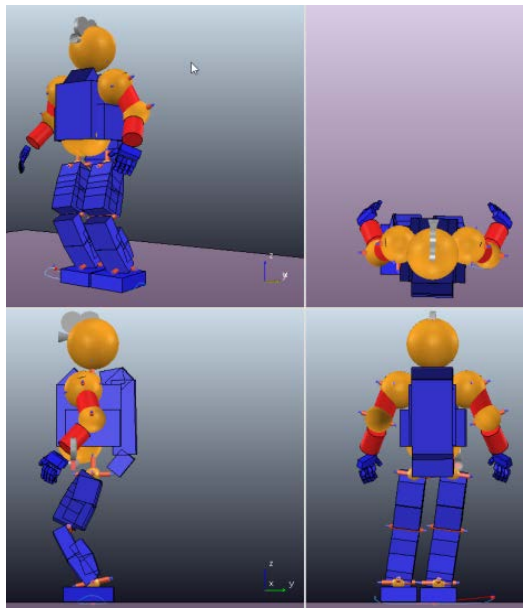

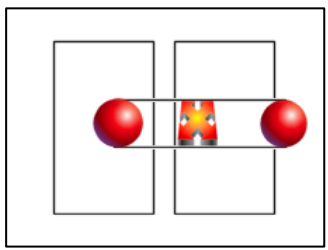

Fig. 8. Desplazamiento lateral

### 3. Primer medio paso de avance

Inicialmente se tienen los dos pies juntos. En ningún momento posterior en la caminata se volverá a tener los pies juntos antes de dar un paso, por lo que este primer paso es especial y en él se avanza la mitad que en los siguientes pasos.

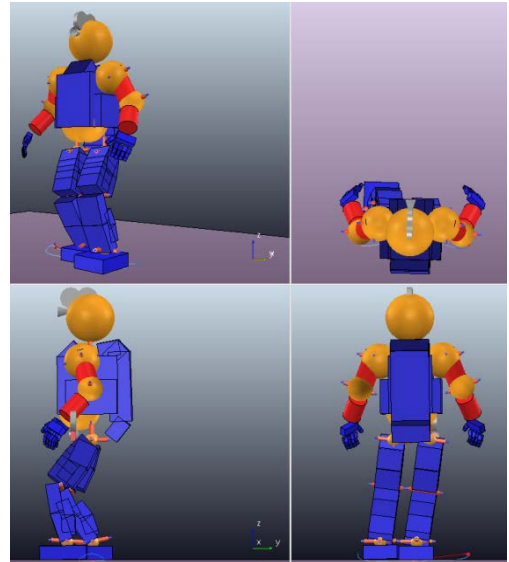

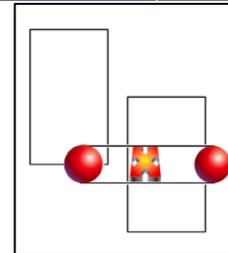

Fig. 9. Primer medio paso de avance

El primer avance corresponde a un medio paso. El pie izquierdo en este caso se levanta y avanza medio paso hasta contactar de nuevo con el suelo. Hasta este momento los ángulos de las articulaciones de las dos piernas serán idénticos, con este primer medio paso desajustamos los ángulos de las piernas, de modo que la evolución cinemática de ambas piernas será distinta desde este momento.

### 4. Desplazamiento lateral

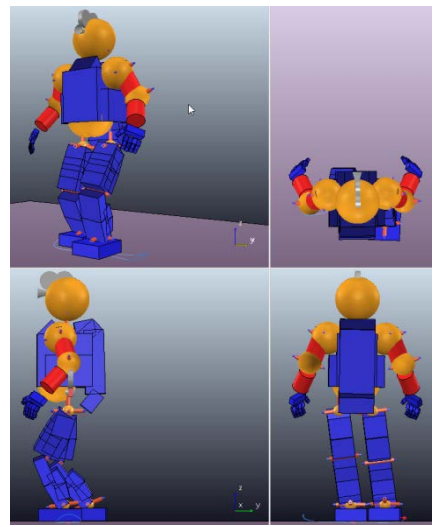

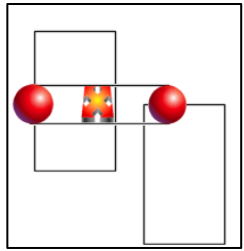

Fig. 10. Movimiento lateral.

El primer paso se ha realizado con la pierna izquierda garantizando que el centro de masas se encuentre encima del pie derecho (en este caso). El segundo paso se realizará con la pierna derecha y se debe garantizar que el centro de masas se encuentre sobre la planta del pie izquierdo.

Para conseguirlo se desplaza, el robot, lateralmente hacia la izquierda, pero este movimiento de desplazamiento no es solo lateral, sino que además el centro de la cadera debe avanzar para situar el centro de masas encima del polígono de soporte generado por el pié derecho.

*5. Segundo paso completo con la pierna derecha*

Una vez que se está de nuevo sobre el polígono de soporte, ahora se desplaza la pierna derecha hasta avanzar por delante de la izquierda.

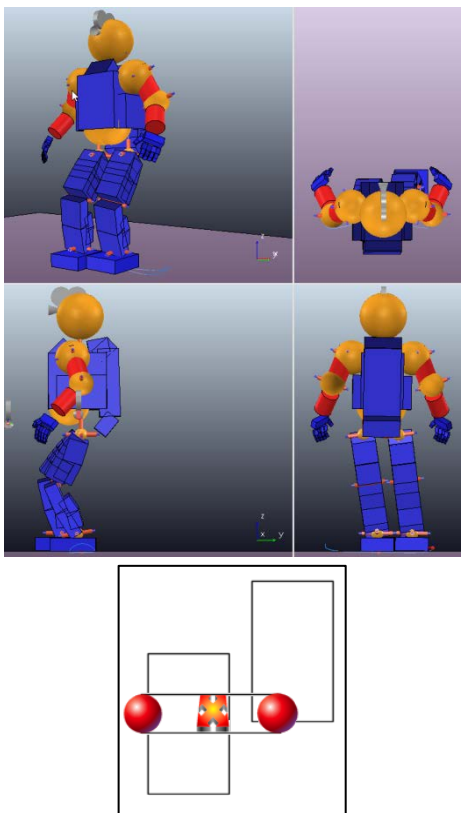

Fig. 11. Segundo paso completo, pierna derecha

Este será el primer paso completo, ya que la pierna derecha se encontraba medio paso detrás de la izquierda y al finalizar la posición se colocará medio paso delante de la pierna izquierda.

Este movimiento se repetirá continuamente durante toda la caminata. Primero se dará un paso con la pierna derecha, nos movemos lateralmente hacia el lado derecho, después se da el paso con la pierna la pierna izquierda, el robot se mueve lateralmente a la izquierda y se comienza de nuevo.

*6. Desplazamiento lateral*

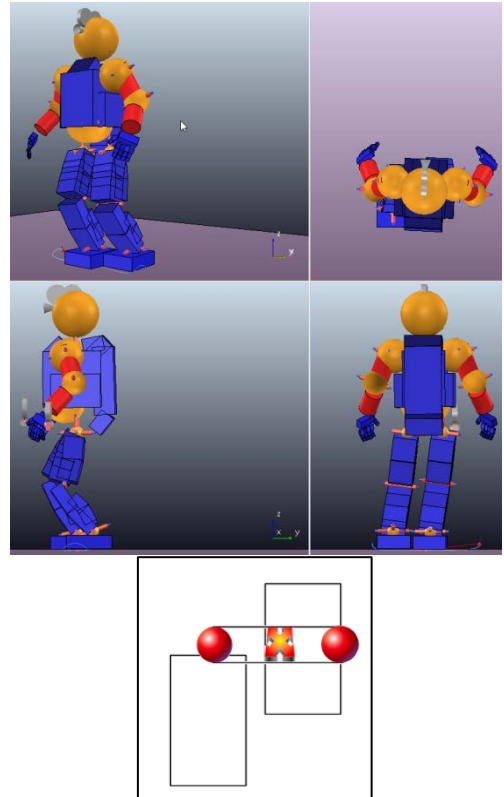

Fig. 12. Desplazamiento lateral para preparar el siguiente paso con la pierna izquierda

Una vez que se da el primer paso completo con la pierna derecha, se debe mover de nuevo el centro de masas para situarlo encima del pie derecho para liberar la pierna contraria para que dé su próximo paso

*7. Paso con la pierna izquierda completo*

Una vez que se sitúa el centro de masas encima de la base del pie derecho, ya se puede dar el paso con la pierna izquierda.

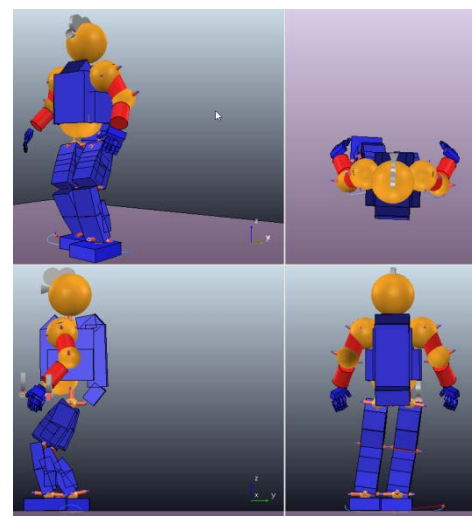

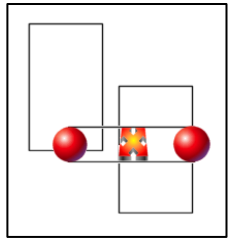

Fig. 13. Avance

A partir de aquí se vuelve a la fase 4 y se repiten éstas fases hasta que se termine la caminata, dando un último medio paso para dejar los dos pies juntos y centrando la cadera.

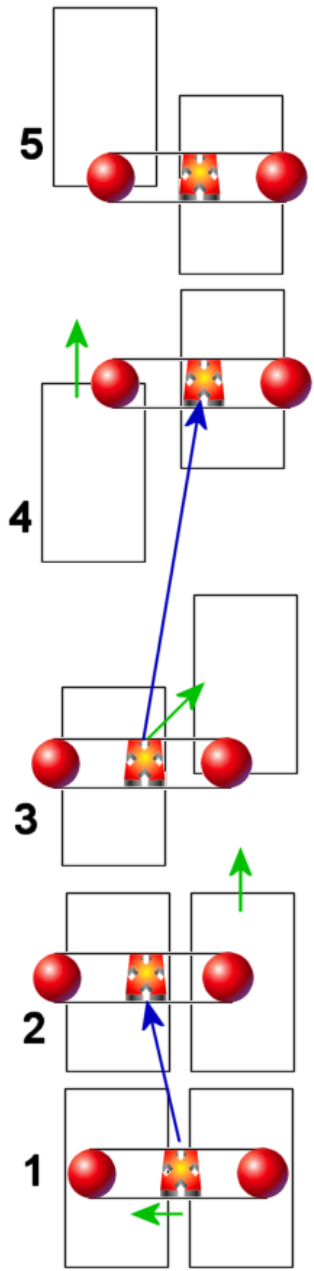

### D. Resolución cinemática

Ahora se sabe, exactamente, los movimientos que se tienen que realizar y las posiciones finales que se tienen que alcanzar en cada movimiento. Con estos datos se trazan las trayectorias que debe seguir el robot.

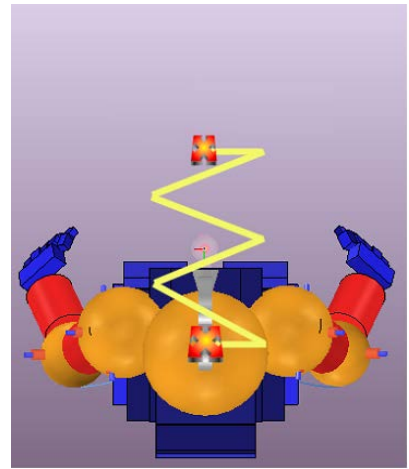

Fig. 14. Trayectoria

En el plano sagital visto desde arriba, se puede ver que el centro de masas debe trazar una trayectoria en zig-zag, visualizada por la línea amarilla.

El robot debe ir balanceando el centro de masas a un lado y a otro para garantizar que el centro de masas se encuentre siempre encima del polígono formado por la base de alguno de los pies.

Ahora que se ve claramente la trayectoria y los movimientos, se debe comenzar solucionando la cinemática del movimiento lateral. Este movimiento permitirá desplazar el centro de masas encima del polígono de soporte formado por el pie que queda en contacto con el suelo.

Rápidamente se ve que el movimiento lateral, siempre que se acepte un ligero desplazamiento hacia arriba y hacia abajo, es trivial, solo se tiene que aplicar "siempre" el mismo ángulo de desplazamiento a ambas piernas en los dos ejes de la cadera y el ángulo inverso en los dos ejes laterales del tobillo.

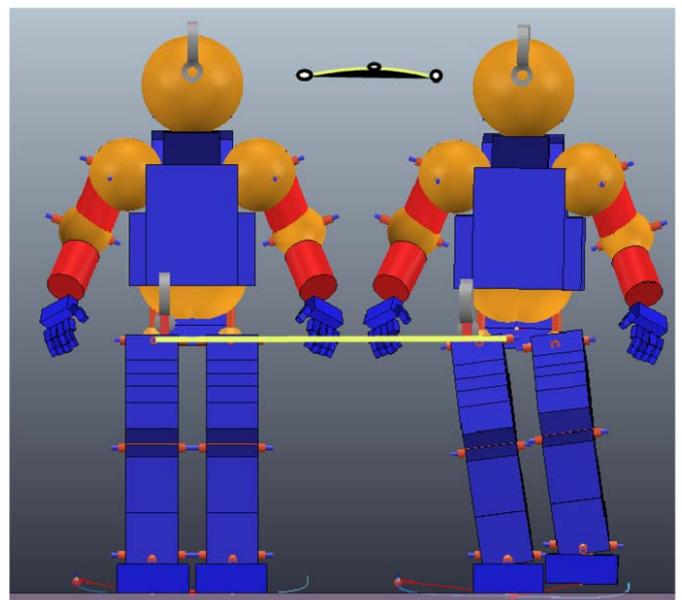

Fig. 15. Lógica del desplazamiento lateral

En la figura 15 se aprecia como al desplazarse lateralmente, el centro de la cadera baja su altura mínimamente. Esto es debido a que el eje de la cadera efectúa una trayectoria circular con respecto al eje del tobillo. Este vaivén es tan asumible que no es necesario corregirlo.

Este sencillo código es el encargado de realizar el desplazamiento lateral:

```
valor = ProgresoTemporal(t, 0, 1, 0.15)
simSetJointTargetPosition(join_cadera_izq_lateral,
valor);
simSetJointTargetPosition(join_cadera_der_lateral,
valor);
simSetJointTargetPosition(join_pie_izq_lateral, -
valor);
simSetJointTargetPosition(join_pie_der_lateral, -
valor);
```

Fig. 16. Código de movimiento

**ProgresoTemporal()** devuelve valores ente 0 y 0.15 distribuidos durante 1 segundo. En cada instante de tiempo devuelve el valor proporcional. Este fragmento de código se ejecuta decenas de veces en un segundo.

Llegado este punto, se debe tener en consideración que el código de simulación tiene un comportamiento muy similar al bucle **void loop()** de Arduino [9]. Arduino es la plataforma más conocida en el mundo para el desarrollo de dispositivos robóticos sencillos. El código principal de un programa para arduino se incluye en la función loop() y esta función repite su ejecución para siempre, cada vez que termina, vuelve a llamarse. En cada iteración, comprobando el instante de tiempo en el que se encuentra, debe calcular los ángulos de todas las articulaciones.

Como ya se tiene solucionado el problema del desplazamiento lateral de modo trivial, ahora se debe conocer la trayectoria que debe seguir el pie para poder dar un paso y una vez que sepamos cómo se tiene que mover el pie tenemos que calcular empleando cinemática inversa los ángulos frontales de la cadera, de la rodilla y del tobillo para que finalmente el pie se sitúe en la posición que se quiere que alcance dentro de la trayectoria.

Para este objetivo se tiene que conseguir una función que defina una trayectoria curva que inicie un ascenso rápido del pie, después debe ascender más lentamente e ir desplazando el pie hacia adelante y finalmente debe bajarlo con una pendiente mayor.

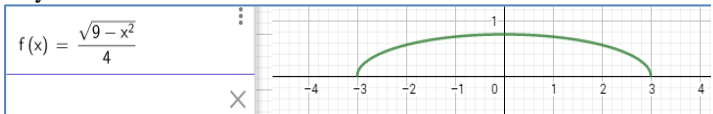

Fig. 17. Función de movimiento del pie.

Finalmente, la función escogida para el recorrido del pie es la que se ver en la Fig. 17 y ésta es la trayectoria que tendrá que seguir la articulación del tobillo para que el robot de un paso.

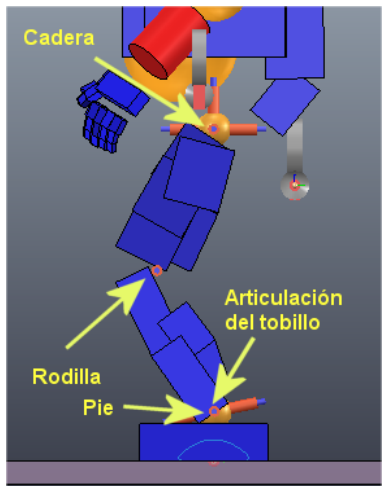

Fig. 18. Articulaciones que se deben gestionar.

En la imagen se puede ver la articulación del tobillo que deberá seguir la trayectoria marcada por la fórmula.

Ahora que ya se tiene la fórmula matemática que permitirá conocer donde debe colocarse la articulación del tobillo en un espacio bidimensional, se debe calcular los ángulos que deben tomar la cadera, la rodilla y al pie para poder alcanzar las distintas posiciones de la trayectoria sin que se desplace la cadera.

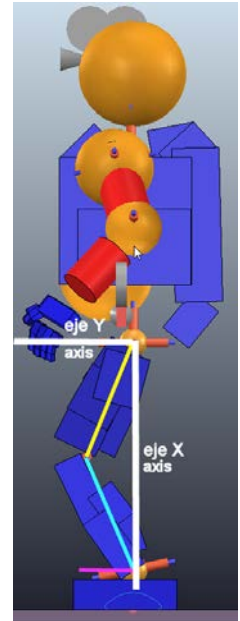

Fig. 19. Partes del triángulo de trabajo.

En este punto es donde entra la trigonometría para ayudarnos.

Inicialmente se puede imaginar que la pierna está formada por tres segmentos, uno asociado a la parte superior de la pierna (amarillo), otro asociado a la parte que se encuentra por debajo de la rodilla (azul) y un último segmento asociado al pie (lila).

Situando estos segmentos en un eje de coordenadas bidimensional se puede encontrar un método trigonométrico para obtener una solución exacta no iterativa.

En la imagen inferior (Fig. 20) se puede ver la reconstrucción del problema cinemático que se tiene que resolver.

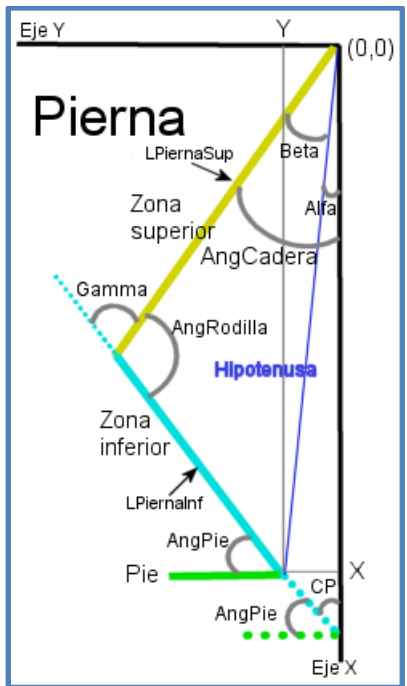

Fig. 20. Problema cinemático.

$$LPiernaSup, LPiernaInf = \text{Longitud de cada tramo}$$
$$Hipotenusa = \sqrt{x^2 + y^2}$$
$$Alfa = atan2(y, x)$$
$$Beta = acos\left(\frac{LPiernaSup^2 - LPiernaInf^2 + Hipotenusa^2}{2 \cdot LPiernaSup \cdot Hipotenusa}\right)$$
$$AngRodilla = Alfa + Beta$$
$$Gamma = acos\left(\frac{LPiernaSup^2 + LPiernaInf^2 - Hipotenusa^2}{2 \cdot LPiernaSup \cdot LongAntBr}\right)$$
$$AngCadera = Gamma - 180$$

Fig. 21. Fórmulas de resolución de la figura 20.

En la pizarra superior (Fig. 21) se pueden ver las fórmulas matemáticas que resuelven el problema y con las se puede obtener los ángulos de la cadera y de la rodilla.

Ahora solo queda resolver el problema del ángulo del pie, pero si se observa detenidamente, al continuar el segmento de la pierna inferior (en azul) se corta con el eje X de coordenadas. Si se calcula la intersección entre el eje X y el segmento de la pierna inferior se tiene el punto de corte. Como consecuencia de ello se tiene un triángulo formado por el punto de intersección, el punto donde debe estar el tobillo y el valor de X en el eje de coordenadas. Conociendo los tres puntos se puede conocer cualquier ángulo del triángulo así que es posible deducir el valor del ángulo **CP** y con este valor el valor de **AngPie = 90º-CP** que es el ángulo que forman el segmento de la pierna inferior y el segmento del pie. Siempre, claro está, asumiendo que el suelo no presenta pendiente.

En este punto se sabe cómo desplazarnos lateralmente y también se sabe cómo dar un paso, ahora solo nos queda saber cómo se puede reposicionar la cadera en diagonal, para posicionar el centro de masas de nuevo sobre el polígono de apoyo del pie, que quedará como soporte en el suelo para que el otro pie dé el siguiente paso.

La componente lateral del desplazamiento diagonal ya se ha explicado y solo queda saber cómo avanza la cadera

En la siguiente imagen se puede ver la situación del robot antes de avanzar la posición de la cadera y la posición final una vez avanzada.

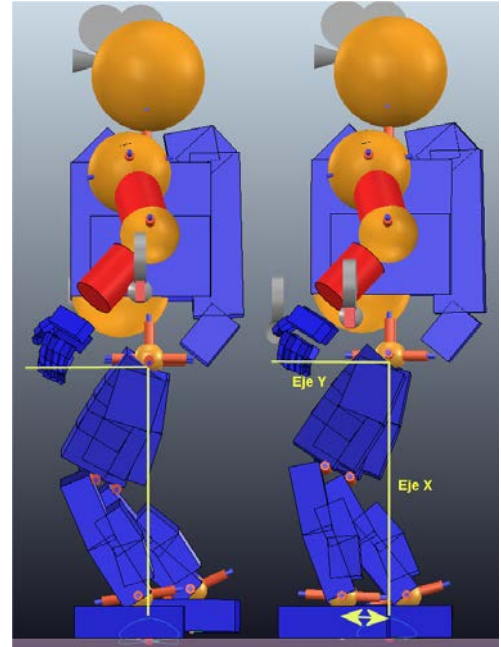

Fig. 22. Eje de cadera.

Si se observa bien, cuando se avanza la cadera, lo que se está haciendo es desplazando el eje de coordenadas hacia adelante o lo que es lo mismo, se está restando el valor del desplazamiento a la coordenada Y, lo que implica que se está restando el desplazamiento realizado al valor de Y, o lo que es lo mismo, se está desplazando el tobillo hacia atrás.

En la figura derecha podemos ver la posición antes de comenzar el desplazamiento y en la figura izquierda vemos la posición después de haber realizado el desplazamiento.

Finalmente, también se incorpora un gráfico en el que se puede ver el triángulo generado por el desplazamiento lateral, con el que se puede calcular la distancia recorrida verticalmente por la cadera como efecto de este desplazamiento.

Aun así, es importante considerar que no resulta útil este ajuste, dado que la distancia es casi despreciable.

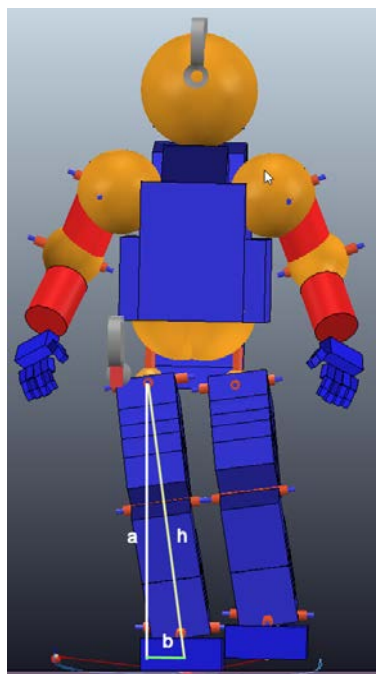

Fig. 23. Desplazamiento lateral.

*E. Conclusiones y trabajo futuro*

Todo el análisis y resolución cinemática propuesta en este trabajo ha sido implementado dentro del simulador de Coppelia Robotics V-REP [10], con lo que hemos podido corroborar la validez de esta línea de resolución geométrica del problema de la caminata bípeda. Todas las imágenes son resultado de las imágenes del robot empleado en la simulación.

En este trabajo hemos conseguido resolver la cinemática estática de un robot bípedo de forma ideal. No hemos tenido en cuenta las inercias generadas por el movimiento, ni tampoco hemos tenido en cuenta la retroalimentación necesaria en estos sistemas de movimiento, debido a que en el mundo real, las posiciones alcanzadas no siempre coinciden con las teóricas. En el siguiente paso, debemos aplicar este estudio teórico a un robot real, en el que nos encontraremos con los problemas inherentes a este mundo, en donde las fórmulas matemáticas devolverán valores cercanos a los reales, pero siempre ligeramente diferentes, lo que ocasionará que nuestro robot real se encuentre con situaciones de pérdida de equilibrio.

*F. Referencias bibliográficas*